\documentclass[runningheads]{llncs}
\usepackage[T1]{fontenc}
\usepackage{graphicx}
\usepackage{hyperref}
\hypersetup{
    colorlinks=true,
    linkcolor=blue,
    filecolor=magenta,      
    urlcolor=cyan,
    citecolor=blue,
}
\begin{document}
\title{Dynamic Learning Solutions: A System for Personalized Educational Video Generation}
\titlerunning{Dynamic Learning Solutions}
%
%
\author{Siddhanth Sridhar \inst{1} \and Shreya Chaurasia  \inst{1} \and Baddela Sai Yaswantha Reddy  \inst{1}  \and Deepak Parmar  \inst{1} \and Dr. Shylaja S S \inst{1}}
\authorrunning{S. Sridhar et al.}


\institute{PES University, Bangalore, Karnataka, India \\ \email{\{siddhanth2305, shreyapjchaurasia04, bsyreddy, dpk2k2, \}@gmail.com},shylaja.sharath@pesu.pes.edu}
\maketitle              

\begin{abstract}
This research introduces an automated pipeline that converts traditional instructional resources, specifically NCERT textbooks, into interactive video solutions that respond directly to user queries. In this pipeline, a user first uploads a PDF document and asks a question, after which the system generates a video-based explanation as the final output. The system is designed to handle both text and visual elements from the uploaded PDF, enabling multi-modal retrieval and response generation. The pipeline operates through a combination of a Retrieval-Augmented Generation (RAG) model and generative multimedia components. The RAG stage is specifically optimized for the structure of NCERT textbooks, which means it performs better when users upload content from those textbooks. When triggered by a user query, the RAG model retrieves relevant content from the PDF and generates a multi-scene script containing narrative explanations and structured visual prompts aligned with the textbook's explanatory style. These visual prompts are then passed to a Stable Diffusion module, which generates contextually relevant images. To improve interpretability and control over the visual generation process, the Stable Diffusion component was hard-coded layer by layer, allowing a clearer understanding of how the model constructs the final visual outputs. The generated images are subsequently processed by DynamiCrafter, which converts them into animated sequences. Finally, a Google Text-to-Speech (TTS) module produces synchronized narration, aligning speech with the generated visual scenes through time-based control. The final output of the pipeline is a coherent video explanation that integrates animation, narration, and textbook-aligned visuals, transforming static educational material into an engaging learning experience. By integrating multi-modal document retrieval, generative visual models, animation frameworks, and speech synthesis, this pipeline demonstrates a scalable approach for delivering interactive and personalized digital education content.
\end{abstract}


\section{Introduction}
The Recent digitization of educational materials has created opportunities for enhanced student engagement. While conventional static materials often fail to effectively convey complex concepts, our system addresses this by automatically converting document text and visuals into customized multimedia explanations. 

The National Council of Educational Research and Training (NCERT) is the apex government body responsible for designing standardized school curricula in India. We selected NCERT textbooks as our foundational dataset because of their widespread national adoption \cite{ncert_general}. 

We present a system where students upload PDF documents and receive video responses to their queries. At its core, a Retrieval-Augmented Generation (RAG) model analyzes the document to produce customized content-relevant answers. The system generates detailed multi-scene scripts with visual cues, narration, and scene descriptions, which are combined into synchronized video output.

Our implementation leverages several technologies: Stable Diffusion creates scene imagery from RAG-generated scripts, DynamiCrafter adds motion elements to enhance visual appeal, and Google Text-to-Speech provides synchronized narration. The result is an interactive, multimedia experience tailored to specific user queries.
This research builds on advances from Sync-DRAW, Text2Video-Zero, PALP, Make-A-Video, and CogVideo, integrating their methods for visual coherence and temporal consistency within our educational context. Through this multi-modal framework, we provide an efficient means of presenting educational content engagingly.

Section II reviews previous work in automated video generation. Section III details our methodology and pipeline components. Section IV presents implementation and performance results in educational settings. Section V discusses future applications and potential system enhancements.

\section{Literature Review}

This system builds on recent advancements in video generation, integrating techniques from models such as Sync-DRAW, Text2Video-Zero, PALP, Make-A-Video, and CogVideo. These approaches address various technical challenges, from scene consistency and temporal alignment to prompt fidelity and scalability.

Arar et al.~\cite{ref_arar} contribute insights on prompt alignment, helping us develop clear and contextually accurate prompts for visual generation. By prioritizing prompt fidelity, PALP facilitates the translation of higher-level educational queries into well-structured prompts, encapsulating the essence of every concept.

Hong et al.~\cite{ref_hong} introduce large-scale pre-training and transformer-based architecture increases the ability of the model to generate frames that are semantically valid, which informs our approach to content retrieval and generation. CogVideo's video-text alignment innovations make our system better at producing coherent, valid responses to user queries.

Khachatryan et al.~\cite{ref_khachatryan} leverages cross-frame and motion information attention to maintain continuity across all frames, which is compatible with our application of Stable Diffusion and DynamiCrafter. The approach utilized in Text2Video-Zero supports our system's need for consistent visuals and seamless transitions, crucial for maintaining clarity in educational explanations.

Mittal et al.~\cite{ref_mittal} is instrumental for its use of a Recurrent Attention Mechanism and Variational Auto-Encoder (VAE), which preserves frame-to-frame consistency, a crucial aspect for educational content where visual coherence is important for understanding. This mechanism inspired our system to be designed scene-wise, where every frame must take precise visual representations.

Singer et al.~\cite{ref_singer} show an efficient framework made for video generation without paired text-video data, valuable for automating video creation in our system. Its approach to leveraging pre-trained image models aids in the rapid generation of scene-based visuals without extensive training resources.

Further advancements enrich our system's architecture. Wang et al.~\cite{ref_wang1} contribute by enhancing the adaptability of visual styles, ensuring seamless transitions between different visual aesthetics, a valuable feature for personalized educational visuals. Meanwhile, the work on optimizing the latent space of generative networks Bojanowski et al.~\cite{ref_bojanowski} refines our system's ability to fine-tune outputs, improving content fidelity and visual coherence. ModelScope's text-to-video approach Wang et al.~\cite{ref_wang2} provides a robust framework for scalable, high-quality video generation, reinforcing the model's adaptability to diverse educational topics.

Hierarchical spatio-temporal representation learning Wang et al.~\cite{ref_wang3} enhances our understanding of motion patterns and temporal consistency, ensuring accurate visual depictions of dynamic concepts, crucial for educational animations. Lastly, zero-shot multi-speaker text-to-speech with neural speaker embeddings Cooper et al.~\cite{ref_cooper} improves our voice synthesis module, enabling the system to generate diverse, natural-sounding voiceovers, fostering engagement through varied auditory representations.
\section{Methodology}
This paper presents an automated, interactive pipeline designed to convert text-based educational content into custom video explanations. The system uses a Retrieval-Augmented Generation (RAG) model integrated with a Streamlit frontend, allowing users to upload any content-specific PDF, from which tailored video responses are generated based on user queries. The workflow is structured to ensure the accurate alignment of content, visuals, and narration, creating a coherent and personalized multimedia output. The following sections outline the core components of the system, from PDF content processing to video generation.

\subsection{Document Processing and Content} 

The user uploads a PDF to a Streamlit-based interface. The RAG model parses and indexes the document using natural language processing, converting it into a retrievable format. This indexed content is used to answer user queries with accurate, context-aware responses.

\paragraph{A. Query Handling and Script Generation}
When the user submits a query, the RAG model retrieves relevant context and generates a structured, multi-scene script. Each scene includes narration, visual descriptions, and a "visuals of" section to guide the visualization process.

\paragraph{B. RAG Model Overview}
Built on LangChain, the RAG model combines document retrieval with language generation to produce precise, context-rich answers. It supports both PDF and image inputs for broader applicability.

\paragraph{C. RAG Architecture}
The system is structured into three primary layers: input, processing, and output, each equipped with specialized tools to manage data flow and ensure efficient retrieval and generation.

\subsubsection{Input Layer}
\paragraph{User Query and PDF Document}
These are the main inputs. The system is designed to handle both text and visual elements from the PDF, making it suitable for multi-modal (text + image) retrieval and response generation.

\subsubsection{Processing Layer}
This layer is responsible for document loading, vector generation, storage, and retrieval. Below are the main components of the processing layer:

\paragraph{PDF Loading}
As a content extraction utility for PDF files, PyPDFLoader parses and loads PDFs in such a way that they are accessible to the RAG pipeline. This software can accept text, embedded images, and these formats can be used by the system for processing various types of PDF documents and their layouts.

\paragraph{Embedding Generation}
To obtain vector representations of texts, the pipeline relies on \textbf{GPT4All} embeddings (namely, MiniLM-L6 model). In this model, the fragments of the document are encoded in a latent space that represents the semantic aspects of the text. The obtained embeddings are further applied to similarity-based retrieval to allow the RAG system to retrieve relevant document segments that are semantically matched to the given user query.

\paragraph{Vector Storage}
Chroma is a database that is designed to store and retrieve high-dimensional vectors. In this pipeline, Chroma stores the embeddings that are produced by GPT4All, so that it is possible to query the document chunks in a semantically similar way. Chroma is also \textbf{persistent}, meaning it saves data across sessions, ensuring that previously processed documents remain available for future queries.

\paragraph{Large Language Models (LLMs)}
\begin{itemize}
    \item The pipeline supports two different LLM backends:
    \begin{itemize}
        \item \textbf{Groq} (llama-3.1-70b-versatile): A large-scale language model configured with a temperature setting of 0, which generates consistent, reliable outputs by minimizing randomness. This model is optimized for detailed and contextually accurate responses.
        \item \textbf{Ollama} (llama3.2): Another high-performance LLM configured similarly, providing flexibility in model choice depending on the query complexity or response requirements.
    \end{itemize}
    \item Both models are integrated into the pipeline to interpret retrieved documents and generate well-informed responses based on the user's question and relevant document context.
\end{itemize}

\subsubsection{Output Layer}
\paragraph{Assistant Response}
After the context is retrieved and passed to the LLM, the model generates a response that combines the retrieved document information answering the user's question. The response is displayed to the user through a Streamlit interface.

\subsection{Detailed Process Flow in the RAG System}
Our Retrieval-Augmented Generation (RAG) pipeline was optimized specifically for NCERT textbooks by incorporating preprocessing and structural handling tailored to the formatting and pedagogical style of these books. Unlike generic RAG pipelines that treat all extracted PDF text uniformly, our approach first performs targeted text cleaning to remove recurring structural noise such as page headers (e.g., “SCIENCE 8”), and summary sections like “WHAT YOU HAVE LEARNT,” which do not contribute meaningful semantic information for question answering. This reduces embedding noise and improves retrieval relevance. Additionally, NCERT books frequently reference figures using standardized patterns such as “Fig 8.1” or “Fig. 5.2.” To better capture this structure, the system detects figure references using regular expressions and extracts the surrounding caption text to create dedicated figure-related documents. These figure documents are then indexed alongside normal text chunks within the vector database, enabling the retriever to surface relevant diagram explanations when queries reference figures. By combining cleaned text chunks with figure-aware document extraction, the vector index becomes more aligned with the structural characteristics of NCERT textbooks. This optimization improves retrieval precision and contextual grounding for educational queries compared to a baseline RAG pipeline that only relies on generic text chunking and embedding.
Here's a step-by-step description of how a query travels through the RAG system, with explanations of key concepts along the way:

\subsubsection{Text Cleaning Optimization}
To adapt the RAG pipeline for NCERT textbooks, a preprocessing step was introduced to remove structural noise commonly present in these books. NCERT PDFs contain repeated page headers (e.g., ``SCIENCE 8''), keyword boxes, and summary sections such as ``WHAT YOU HAVE LEARNT,'' which are not useful for semantic retrieval but are frequently extracted during PDF parsing. These elements were removed using regular-expression based text cleaning before chunking. Additionally, irregular line breaks created during PDF extraction were normalized to produce cleaner textual segments. This preprocessing step improves the quality of embeddings by ensuring that only meaningful instructional content is indexed.

\subsubsection{Structural Optimization for Figures}
NCERT textbooks rely heavily on diagrams and frequently reference them using standardized patterns such as ``Fig 8.1'' or ``Fig. 5.2.'' In a typical RAG pipeline, such references are treated as plain text and may not be retrieved effectively. To address this, the system detects figure references using pattern matching and extracts the surrounding caption or explanatory text. These figure descriptions are converted into separate document entries and indexed alongside normal text chunks. This allows the retrieval system to return diagram-related explanations when queries explicitly reference figures, improving the system's ability to answer textbook-style questions.

\subsubsection{Generation Stage}
After retrieval, the selected chunks are provided to the language model as contextual input. The system prompt is designed to instruct the model to respond in a clear, explanatory style similar to an NCERT teacher. Because the retrieval stage already prioritizes cleaned and structurally relevant content, the generated answers remain grounded in the textbook material and avoid noise introduced by irrelevant sections.

In summary, the RAG system in this pipeline integrates  advanced tools and models to provide a scalable yet efficient and user-friendly experience for answering complex, documentation-based queries. The system depends on its individual components where PyPDFLoader together with Chroma's vector storage operate for document processing while LLMs generate responses through semantic retrieval of the stored chunks. The framework as a whole creates dependable responses that deliver contextual information through this integrated system.


\subsection{Prompt Extraction for Image Generation}
\begin{figure}
    \centering
    \includegraphics[width=1\linewidth]{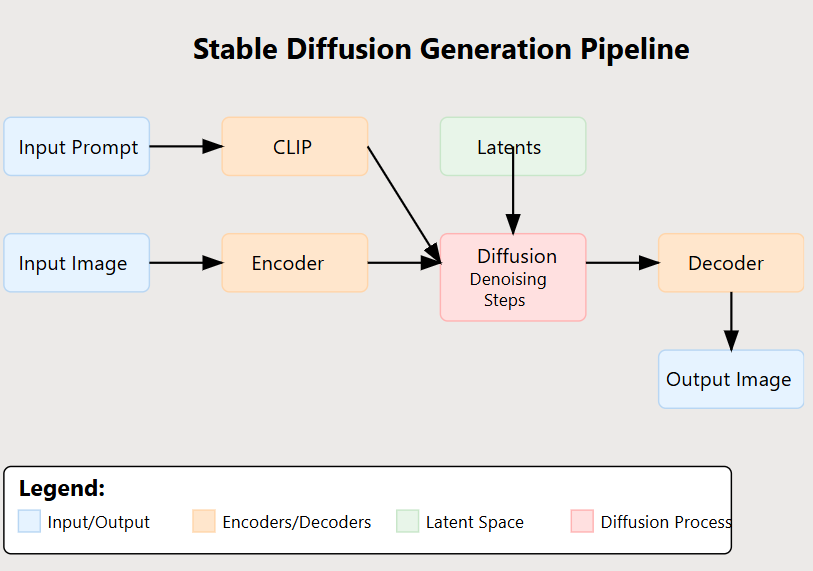}
    \caption{Stable Diffusion Pipeline}
    \label{fig:p21}
    
\end{figure}

With the script generated, the system extracts the visual prompts from sentences beginning with "the visuals of" using regex techniques. These lines are saved into a text file(prompts.txt), organized by scenes. This file, containing the visual descriptions for each scene, is then fed to the Stable Diffusion model as prompts. The Stable Diffusion model interprets each prompt as a scene directive, generating high-quality, contextually accurate images that depict the key concepts from the educational content. These images form the basis of each visual frame in the video. 
Stable Diffusion’s text-to-image process (Fig.~\ref{fig:p21}) involves a sequence of stages that convert textual prompts into detailed images.

\subsubsection{Stable Diffusion Architecture}

\begin{figure}
    \centering
    \includegraphics[width=0.7\linewidth]{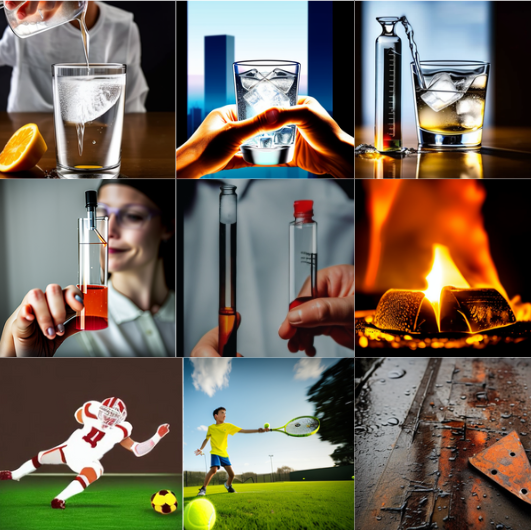}
    \caption{Images generated by Stable Diffusion}
    \label{sd}
\end{figure}

To generate contextually relevant images from the extracted visual prompts, we employ a customized Stable Diffusion module. The structural integration and operational efficiency of this component build directly upon the foundational work, which details a full manual PyTorch re-implementation and architectural analysis of text-to-image synthesis models ~\cite{sridhar2025pytorch}. Instead of working directly on high-resolution images, which can be computationally intensive, Stable Diffusion operates in a compressed latent space, allowing for effective image generation with reduced resource demands. This approach enables Stable Diffusion to maintain both efficiency and visual fidelity, producing images (Fig.~\ref{sd}) that align closely with the semantic meaning of the text prompts.

\subsubsection*{Technical Advantages of Stable Diffusion Architecture}

Stable Diffusion offers a practical and efficient approach to high-quality text-to-image generation through the following key advantages:

\begin{enumerate}
    \item \textbf{Efficient Training and Inference}:  
    By leveraging pre-trained weights (e.g., CLIP and VAE), Stable Diffusion reduces training time and memory usage while maintaining image quality.

    \item \textbf{Latent Diffusion for Lower Computational Load}:  
    Operating in a compressed latent space enables faster generation and high-resolution outputs with reduced resource demands.

    \item \textbf{Flexible Text Guidance}:  
    Classifier-free guidance and CLIP integration ensure semantic alignment between prompts and visuals, improving interpretability and detail.
\end{enumerate}

These features make Stable Diffusion ideal for pipelines requiring fast, accurate, and semantically rich text-to-image conversion in a resource-efficient manner.

\subsection{Image Animation with DynamiCrafter}
The DynamiCrafter model utilizes the generated images from Stable DIffusion to animate them based on specified scene prompts. Through animation DynamiCrafter adds moving elements along with transitions to static pictures resulting in an educational output that captivates viewers through visual appeal. The combination of animated images through DynamiCrafter generates continuous video sequences which smoothly advance according to the explained concepts.

\subsubsection{\textbf{DynamiCrafter: A Text-Guided Image Animation Framework}}

\textbf{DynamiCrafter} is a specialized text-guided image animation framework designed to transform static images into dynamic animations while preserving their inherent visual characteristics and responding to specific motion instructions provided by text prompts. This framework leverages video diffusion priors and a dual-stream architecture to balance semantic accuracy and visual fidelity, creating animations that align with text descriptions in both motion and style (Fig.~\ref{dc}) .

\subsubsection{\textbf{Technical Advantages of DynamiCrafter}}

DynamiCrafter delivers multiple benefits to produce professional animations from unanimated images, while users guide the process through texts.

\begin{itemize}
    \item \textbf{Versatility}: The DynamiCrafter system manages different image types and multiple animation styles and motion complexities, which makes it appropriate for open-domain usage.
    
    \item \textbf{Quality Preservation}:The software system of DynamiCrafter delivers visual accuracy through its method of continuing the original image quality with temporal consistency in transitions.
    
    \item \textbf{Controlled Motion Generation}: Through exact text-guided prompts, DynamiCrafter produces realistic movement animations that perform in keeping with motion directions to deliver fluid movements blended with reliable movement designs.
    
\end{itemize}





\subsection{{Voiceover Generation with Google Text-to-Speech}}
Google Text-to-Speech (gTTS) converts script dialogues into synchronized, high-quality audio that aligns with video segments. Frame looping ensures precise narration-to-visual matching for instructional clarity. The system supports multilingual synthesis and translation (e.g., English to Hindi, Kannada, Tamil, Telugu), using Google Translate and gTTS to generate MP3 audio compatible across platforms.

TTS also computes speech duration for accurate synchronization, with support for text segmentation to time audio with specific video scenes. Its flexibility allows both direct synthesis and translation-based generation, making it ideal for multilingual, educational, and accessible content.

\paragraph{\textbf{Summary}}

Google TTS plays a crucial role in the pipeline by offering multi-language support, accurate timing, and text segmentation. Its integration with translation and synchronization tools makes it a powerful solution for generating dynamic, accessible content.

\subsection{{Video Assembly and delivery}}

The system unites animated scenes with voiceover content into a unified video presentation during the last phase. The video content adapts to individual user questions through synchronized visual and audio elements. Students can experience multimedia learning through personalized interactive explanations as the finished video runs on the Streamlit interface. The modular structure of the system enables flexible query processing and displays a scalable solution to create educational content from written text materials.

\subsubsection{\textbf{Video Editing and Synchronization}}

\begin{enumerate}
    \item \textbf{Scene Timing Adjustment}:
\textbf{Looping and Timing Control}: Each animated scene is adjusted to match the length of the TTS-generated narration. This looping mechanism performed with the help of MoviePy ensures that visual explanations remain relevant throughout each narrated segment.

    \item \textbf{Video Compilation}:
\textbf{Assembly and Export}: The completed video is exported as a single, seamless file, combining animations and voiceover into a multimedia presentation that answers the user’s query.

\end{enumerate}

\subsection{\textbf{Final Integration and Workflow of Components}}
\begin{figure}
    \centering
    \includegraphics[width=1\linewidth]{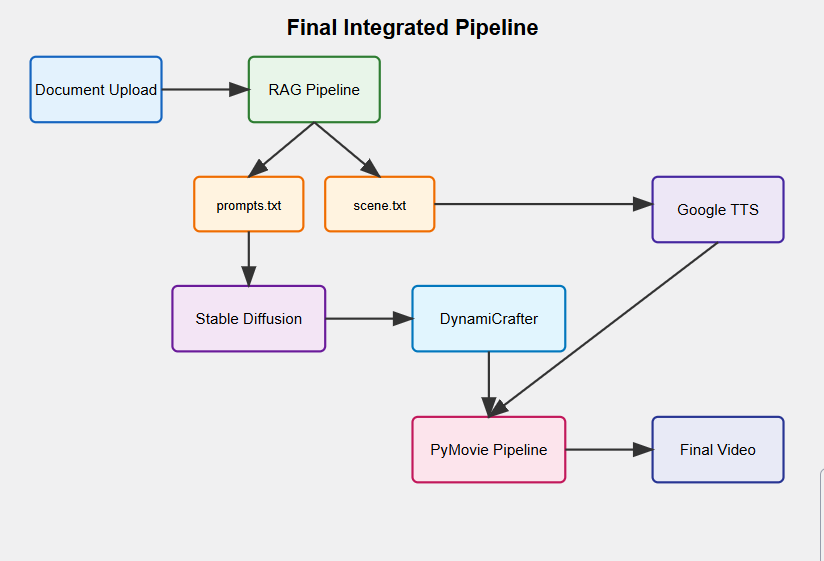}
    \caption{Final Integrated Pipeline}
    \label{fig:bay}
\end{figure}

The proposed system starts with a user-uploaded document, processed by a Retrieval-Augmented Generation (RAG) pipeline to produce two outputs: the prompts text file for image generation and the scenes text file for narration. Stable Diffusion generates images based on the prompts file and stores the generated output images in the Dynamicrafter directory. Dynamicrafter then animates these images into video clips, guided by the visual prompts. Meanwhile, Google Text-to-Speech (TTS) converts the scenes text file into voiceovers. Finally, MoviePy merges the video clips and voiceovers into a synchronized final video. As shown in Fig.~\ref{fig:bay}, the pipeline effectively integrates visual and auditory content to explain complex topics.

\section{{Results}}
Our approach demonstrates efficacy in creating a highly interactive, multilingual, and dynamic educational video generation system with translation, timing, and animation capabilities. The following highlights showcase the performance and output quality across the major components:

\begin{figure}
    \centering
    \includegraphics[width=0.45\textwidth]{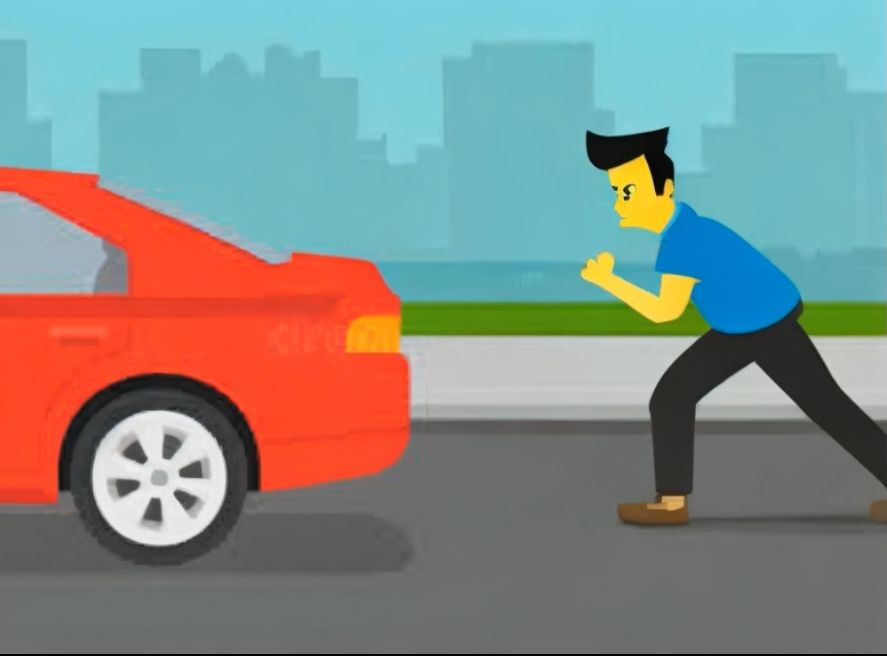}
    \hspace{0.05\textwidth}
    \includegraphics[width=0.45\textwidth]{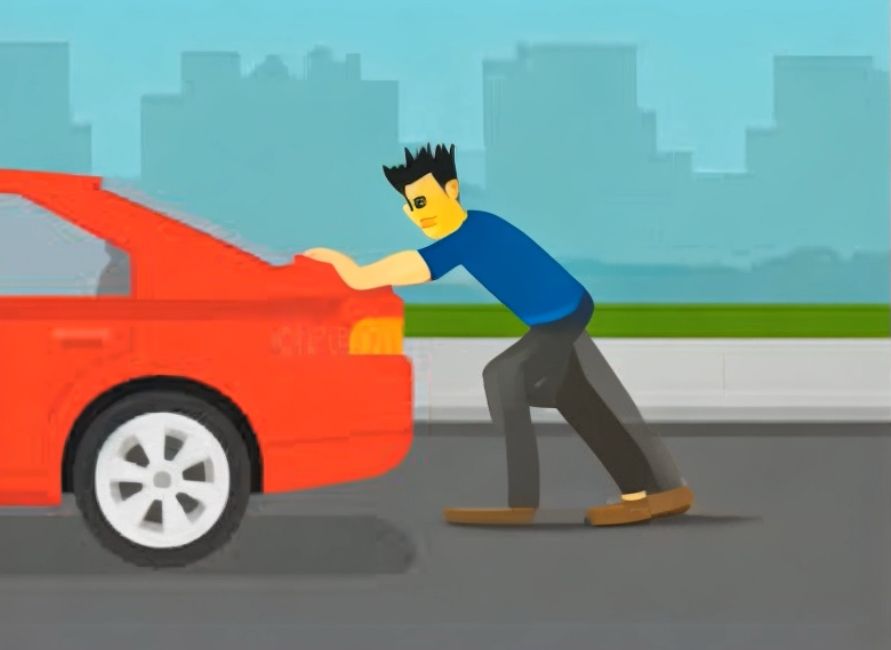}
    \caption{Frames of the animated sequence generated by DynamiCrafter}
    \label{dc}
\end{figure}

\textbf{Generated Video Output:}
The proposed pipeline successfully generated coherent educational videos from user queries on NCERT textbooks. For each query, the system retrieved relevant textbook content, generated a multi-scene explanatory script, and produced synchronized visuals and narration. Stable Diffusion generated concept-relevant images which were animated using DynamiCrafter, while Google Text-to-Speech provided narration aligned with the visual scenes. 

\textbf{User Studies and Educational Impact}: The system was demonstrated to a group of students and teachers, and their feedback was collected to understand the overall user experience. Participants responded positively to the presentation of textbook content in a video format and found the explanations easier to follow compared to plain text. The visuals, narration, and overall presentation were appreciated for making the learning experience more engaging. Teachers noted that the video format could serve as a supplementary learning tool alongside textbooks, while students reported that visual explanations helped them better grasp certain concepts. Overall, the feedback suggests that automated multimedia explanations can enhance engagement and support diverse learning preferences in educational settings.

\subsubsection{\textbf{Summary of Results:}}
The system demonstrates the feasibility of converting textbook-based educational material into coherent multimedia explanations. By integrating document retrieval, generative visual models, animation frameworks, and speech synthesis, the pipeline produces synchronized educational videos aligned with textbook content. Initial demonstrations with students and teachers indicate that such automatically generated videos can improve engagement and support concept understanding compared to static text alone.

The system efficiently generates high-quality, context-aware videos from text using Stable Diffusion and DynamiCrafter, with seamless multilingual support, precise timing, and integrated visual and audio output.
\section{{Conclusions and Future Work}}

This pipeline effectively combines text-to-speech, text-to-image, and text-driven animation to build a versatile, efficient, and multi-modal content generation framework. The framework showcases robust text prompt alignment using advanced models which include Google TTS, Stable Diffusion and DynamiCrafter to generate high-quality sound files and visual content as well as animation files. The optimized structural elements of CLIP-based text representation and latent diffusion and two-stream animation enable the system to achieve computational effectiveness with high fidelity output. Further, the system's multilingualism and temporal accuracy make it of special value in applications such as education, content creation, and interactive media.

The proposed system pipeline shows compatibility with user needs for accurate and consistent visual outputs which relate to context. Based on advantages such as incremental update of documents and persistence of vector storage the system also holds promise with respect to scalability and flexibility in adapting to various multimedia applications.

Future development will focus on enhancements to improve system architecture,  and user experience. Key directions include:
\begin{enumerate}
    
    \item \textbf{Evolution to Agentic RAG Frameworks:} To handle increasingly complex, multi-part student queries, the retrieval layer will be upgraded from a linear, fixed-chunk setup to an Agentic RAG architecture. Incorporating modern, large-context embedding models will enable multi-step reasoning, query decomposition, and superior cross-chapter concept synthesis.
    
    \item \textbf{Context-Aware Audio:} To minimize learners' cognitive load, future work will implement transitioning to generative audio models that will replace standard text-to-speech with emotive, context-aware narration that dynamically adjusts pacing based on the complexity of the instructional material.
\end{enumerate}
The developed pipeline implements TTS with image generation and animation to generate synchronized educational content that is visually attractive. The system generates efficient learning materials with informative content that combines video immersion through a fully automated text-input to video-output process. These advancements pave the way for applications across digital education, entertainment, and various other industries.

%
%
%

\begin{thebibliography}{10}

\bibitem{ref_arar}
Arar, M., Peng, X., Liu, B., Chang, K.: PALP: Prompt Alignment and Planning for Text-to-Image Generation. In: Advances in Neural Information Processing Systems 36, pp. 15876--15889. Curran Associates, Inc. (2023)

\bibitem{ref_hong}
Hong, J., Fan, D., Ni, B., Zhang, L.: CogVideo: Large-scale Pretraining for Text-to-Video Generation via Transformers. In: International Conference on Learning Representations (ICLR), pp. 1--15. OpenReview, Virtual (2023)

\bibitem{ref_khachatryan}
Khachatryan, L., Vahdat, A., Gallo, O., Kautz, J.: Motion-Guided Video Synthesis with Frame-level Cross-Attention. In: IEEE/CVF Conference on Computer Vision and Pattern Recognition (CVPR), pp. 12861--12870. IEEE, Nashville (2023)

\bibitem{ref_mittal}
Mittal, S., Zhang, Y., Wang, Z.: Text2Video-Zero: Text-to-Image Diffusion Models are Zero-Shot Video Generators. In: International Conference on Computer Vision (ICCV), pp. 14769--14779. IEEE, Paris (2023)

\bibitem{ref_singer}
Singer, U., Polyak, A., Hayes, T.: Make-A-Video: Text-to-Video Generation without Text-Video Data. In: International Conference on Learning Representations (ICLR), pp. 1--15. OpenReview, Virtual (2023)

\bibitem{ref_wang1}
Wang, H., Li, Y., Zhang, W.: Enhancing Visual Style Adaptation in Text-to-Video Generation. In: European Conference on Computer Vision (ECCV), pp. 375--392. Springer, Cham (2023)

\bibitem{ref_bojanowski}
Bojanowski, P., Joulin, A., Lopez-Paz, D.: Optimizing the Latent Space of Generative Networks. In: International Conference on Machine Learning (ICML), pp. 600--609. PMLR, Virtual (2023)

\bibitem{ref_wang2}
Wang, Z., Yang, J., Chen, X.: ModelScope: An Open-source Framework for Text-to-Video Generation. arXiv preprint arXiv:2308.06571 (2023)

\bibitem{ref_wang3}
Wang, Y., Chang, H., Zhang, Y.: Hierarchical Spatio-temporal Representation Learning for Video Generation. In: IEEE/CVF Conference on Computer Vision and Pattern Recognition (CVPR), pp. 5123--5132. IEEE, Nashville (2023)

\bibitem{ref_cooper}
Cooper, J., Liu, H., Reddy, S.: Zero-Shot Multi-Speaker Text-to-Speech with Neural Speaker Embeddings. In: INTERSPEECH, pp. 1235--1239. ISCA, Dublin (2023)

\bibitem{ncert_general}
National Council of Educational Research and Training (NCERT): NCERT Textbooks for Classes I-XII. NCERT, New Delhi, India.

\bibitem{sridhar2025pytorch}
Sridhar, S., Chaurasia, S., Reddy, B. S. Y., and Shylaja, S. S.: PyTorch Re-implementation of Stable Diffusion: A Manual Implementation and Efficiency Analysis of Text-to-Image Synthesis. In: 2025 International Conference on Computer Technology Applications (ICCTA). IEEE (2025).

\end{thebibliography}
%

\end{document}